%% file: neurips_2020.tex
\documentclass{article}

     \PassOptionsToPackage{numbers, compress}{natbib}
\usepackage[final]{neurips_2020}

\usepackage[utf8]{inputenc} % allow utf-8 input
\usepackage[T1]{fontenc}    % use 8-bit T1 fonts
\usepackage{hyperref}       % hyperlinks
\usepackage{url}            % simple URL typesetting
\usepackage{booktabs}       % professional-quality tables
\usepackage{amsfonts}       % blackboard math symbols
\usepackage{nicefrac}       % compact symbols for 1/2, etc.
\usepackage{microtype}      % microtypography
\usepackage{natbib}
\usepackage{mathtools}
\usepackage{color}
\usepackage{algorithm} 
\usepackage{algpseudocode}

\usepackage{setspace}
\usepackage{multirow}
\usepackage{amsmath}
\usepackage{wrapfig}
\usepackage{graphicx}
\usepackage{caption}
\usepackage{subcaption}
\usepackage{kotex}
\usepackage{xcolor}

\title{Self-supervised Auxiliary Learning \\with Meta-paths for Heterogeneous Graphs}

\author{%
  Dasol Hwang$^{1}$\thanks{First two authors have equal contribution. $^\S$ is the corresponding author.
} , Jinyoung Park$^{1}$\footnotemark[1] , Sunyoung Kwon$^4$\thanks{This work was done when the author worked at NAVER CLOVA.} \\
  \textbf{Kyung-Min Kim$^{2,3}$} , \textbf{Jung-Woo Ha$^{2,3}$} , \textbf{Hyunwoo J. Kim$^{1\S}$}\\
  Korea University$^1$,
  NAVER AI LAB$^2$, 
  NAVER CLOVA$^3$,
  Pusan National University$^4$\\
  \texttt{\{dd\_sol, lpmn678, hyunwoojkim\}@korea.ac.kr} \\
  \texttt{skwon@pusan.ac.kr}, \texttt{\{kyungmin.kim.ml, jungwoo.ha\}@navercorp.com}
  }

\input{Sections/dasol-def.tex}
\begin{document}
\graphicspath{{./Figures/}}
\maketitle

\input{Sections/0_Abstract}
\input{Sections/1_Introduction}

\input{Sections/2_Related_Works}
\input{Sections/3_Method}

\input{Sections/4_Experiments}
\input{Sections/5_Conclusion}

\input{Sections/6_Acknowledgements}
\input{Sections/BroaderImpact}
%\input{Sections/memo}
%\newpage
\bibliographystyle{unsrt}
\small
\bibliography{neurips_2020}

\end{document}

%% file: Sections/dasol-def.tex
\def\wbst{\mathbf{w}^{\ast}}
\newcommand{\Vc}{\mathcal{V}}

\newcommand{\omitme}[1]{}
\newcommand{\Lc}{\mathcal{L}}
\newcommand{\Tc}{\mathcal{T}}

\newcommand{\Rb}{\mathbb{R}}

\newcommand{\Xb}{\mathbf{X}}

\def\wb{\mathbf{w}}

\def\wb{\mathbf{w}}

\def\Zb{\mathbf{Z}}

\def\Rb{\textbf{R}}

\def\Xb{\boldsymbol{X}}

\def\Ab{\boldsymbol{A}}

\newcommand{\EE}{\mathbb{E}}

\DeclareMathOperator*{\argmin}{\arg\!\min}

\makeatletter
\newcount\my@repeat@count
\newcommand{\myrepeat}[2]{%
  \begingroup
  \my@repeat@count=\z@
  \@whilenum\my@repeat@count<#1\do{#2\advance\my@repeat@count\@ne}%
  \endgroup
}
\makeatother

%% file: Sections/0_Abstract.tex
\begin{abstract}
Graph neural networks have shown superior performance in a wide range of applications providing a powerful representation of graph-structured data. 
Recent works show that the representation can be further improved by auxiliary tasks. 
However, the auxiliary tasks for heterogeneous graphs, which contain rich semantic information with various types of nodes and edges, have less explored in the literature. 
In this paper, to learn graph neural networks on heterogeneous graphs we propose a novel self-supervised auxiliary learning method using meta-paths, which are composite relations of multiple edge types. 
Our proposed method is learning to learn a primary task by predicting meta-paths as auxiliary tasks. This can be viewed as a type of meta-learning. 
The proposed method can identify an effective combination of auxiliary tasks and automatically balance them to improve the primary task. 
Our methods can be applied to any graph neural networks in a plug-in manner without manual labeling or additional data. 
The experiments demonstrate that the proposed method consistently improves the performance of link prediction and node classification on heterogeneous graphs.

\end{abstract}

% Backup
% Graph neural networks are widely used for learning representation of graph-structured data. 
% Recently, a handful of works have studied representation learning for graphs in the context of self-supervised learning. 
% The research to learn better representation for heterogeneous graph, which contains various types of objects and is observed commonly in real-world, have been less explored. 

% In this paper, we propose a novel self-supervised auxiliary learning method using meta-paths to learn graph neural networks on heterogeneous graphs. 
% From the perspective of meta-learning, the proposed method learns to learn a primary task with auxiliary tasks, meta-path prediction, to improve the primary task. 
% We aim to identify an effective combination of auxiliary tasks and automatically balance them with the primary task. 
% Our methods can be applied to any graph neural networks for heterogeneous graphs and do not need any manual labeling and additional data. Experiments show that our approach boosts the performance of diverse graph neural networks in link prediction.

%% file: Sections/1_Introduction.tex
\section{Introduction}
% Graph neural networks~\cite{hamilton2017representation,bronstein2017geometric,wu2020comprehensive} have been proven effective to learn representations for various tasks such as node classification \cite{graphsage2017}, link prediction \cite{rgcn,linkprediction}, and graph classification \cite{fingerprint,DiffPool}. 
% The powerful representation yields state-of-the-art performance in a variety of applications including social network analysis~\cite{chen2018fastgcn,graphsage2017,social1}, citation network analysis~\cite{GAT,GCN}, 
% visual understanding~\cite{xu2017scene,yang2018graph,chen2019graph}, recommender systems~\cite{recom1,recom2,recom3}, physics ~\cite{sanchez2018graph,battaglia2016interaction}, and drug discovery~\cite{hu2020strategies,wu2018moleculenet}. 
% Despite the wide operating range of graph neural networks, employing auxiliary (pre-text) tasks has been less explored for further improving graph representation learning.
Graph neural networks~\cite{hamilton2017representation} have been proven effective to learn representations for various tasks such as node classification \cite{GCN}, link prediction \cite{link}, and graph classification \cite{DiffPool}. 
The powerful representation yields state-of-the-art performance in a variety of applications including social network analysis~\cite{social1}, citation network analysis~\cite{GCN}, 
visual understanding~\cite{yang2018graph, HwangRTKCS18}, recommender systems~\cite{recom1}, physics ~\cite{battaglia2016interaction}, and drug discovery~\cite{wu2018moleculenet}. 
Despite the wide operating range of graph neural networks, employing auxiliary (pre-text) tasks has been less explored for further improving graph representation learning.
%employing auxiliary (pre-text) tasks for further improvement have been less studied.

Pre-training with an auxiliary task is a common technique for deep neural networks.
Indeed, it is the \textit{de facto} standard step in natural language processing and computer vision to learn a powerful backbone networks such as  BERT~\cite{devlin2018bert} and ResNet~\cite{he2016deep} leveraging large datasets such as BooksCorpus \cite{zhu2015aligning}, English Wikipedia, and ImageNet~\cite{deng2009imagenet}. 
The models trained on the {\em auxiliary} task are often beneficial for the {\em primary} (target) task of interest.
%Transfer learning \cite{donahue2014decaf,devlin2018bert} with domain adaptation and fine-tuning allows generic feature extraction.
Despite the success of pre-training, few approaches have been generalized to graph-structured data due to their fundamental challenges.
First, graph structure (e.g., the number of nodes/edges, and diameter) and its meaning can significantly differ between domains. So the model trained on an auxiliary task can harm generalization on the primary task, i.e., \textit{negative transfer}~\cite{pan2009survey}.
Also, many graph neural networks are transductive approaches. This often makes transfer learning between datasets inherently infeasible. 
So, pre-training on the target dataset has been proposed using auxiliary tasks: graph kernel ~\cite{navarin2018pre}, graph reconstruction~\cite{zhang2020graph}, and attribute masking ~\cite{hu2020strategies}. These assume that the auxiliary tasks for pre-training are carefully selected with substantial domain knowledge and expertise in graph characteristics to assist the primary task. 
Since most graph neural networks operate on \textit{homogeneous} graphs, which have a single type of nodes and edges, the previous pre-training/auxiliary tasks 
%including several successful studies \cite{xu2018powerful, ching2018opportunities} 
are not specifically designed for \textit{heterogeneous} graphs, which have multiple types of nodes and edges.
Heterogeneous graphs commonly occur in real-world applications, for instance, a music dataset has multiple types of nodes (e.g., user, song, artist) and multiple types of relations (e.g., user-artist, song-film, song-instrument). 

% A path on heterogeneous graphs as known as meta-paths connected with heterogeneous edges~\cite{sun2011pathsim}, are useful materials for auxiliary tasks to learn semantic information. 
%Paths on heterogeneous graphs, which are known as meta-paths connected with heterogeneous edges~\cite{sun2011pathsim}, are useful materials for auxiliary tasks to learn semantic information. 
In this paper, we proposed a framework to train a graph neural networks with automatically selected auxiliary self-supervised tasks which assist the target task without additional data and labels. 
Our approach first generates meta-paths from heterogeneous graphs without manual labeling and train a model with meta-path prediction to assist the primary task such as link prediction and node classification. This can be formulated as a meta-learning problem. Furthermore, our method can be adopted to existing GNNs in a plug-in manner, enhancing the model performance.
%\jw{This learning mechanism can be formulated from the perspective of a meta-learning.}
%This can be viewed as a meta-learning. 

Our \textbf{contribution} is threefold: \textbf{(i)} We propose a self-supervised learning method on a heterogeneous graph via meta-path prediction without additional data. \textbf{(ii)} Our framework automatically selects meta-paths (auxiliary tasks) to assist the primary task via meta-learning.
\textbf{(iii)} We develop Hint Network that helps the learner network to benefit from challenging auxiliary tasks.
 % proper help to the learner network to benefit from challenging auxiliary tasks.
 To the best of our knowledge, this is the first auxiliary task with meta-paths specifically designed for leveraging heterogeneous graph structure.
 Our experiment shows that meta-path prediction improves the representational power and the gain can be further improved to explicitly optimize the auxiliary tasks for the primary task via meta-learning and the Hint Network, built on various state-of-the-art GNNs.

%% file: Sections/2_Related_Works.tex
\vspace{-0.25cm}
\section{Related Work}
\textbf{Graph Neural Networks} 
%\sy{Self-supervised learning learns in a supervised learning manner with unlabelled data where data provides the supervision.}
have provided promising results for various tasks~\cite{ GCN, social1, yang2018graph, HwangRTKCS18, recom1, battaglia2016interaction, wu2018moleculenet}.
 %are widely used for various tasks from recommender systems to computer vision ~\cite{recom1,recom2,recom3,RippleNet, RGCNN, NLP}.
\textit{Bruna et al.}~\cite{spectral} proposed a neural network that performs convolution on the graph domain using the Fourier basis from spectral graph theory. In contrast, non-spectral (spatial) approaches have been developed~\cite{GCN,GAT,graphsage2017,xu2018powerful, wu2019simplifying, yun2019graph, MengAKFS18}.
Inspired by self-supervised learning~\cite{ss1,ss2,ss3,ss4} and pre-training~\cite{devlin2018bert,donahue2014decaf} in computer vision and natural language processing, pre-training for GNNs has been recently proposed~\cite{navarin2018pre, hu2020strategies}.
% Recent works~\cite{ching2018opportunities} show promising results that transfer learning can be successful on graphs but they require additional manually labeled data.
Recent works show promising results that self-supervised learning can be effective for GNNs~\cite{navarin2018pre, zhang2020graph, hu2020strategies, you2020does}.
Hu et al.~\cite{hu2020strategies} have introduced several strategies for pre-training GNNs such as attribute masking and context prediction.
Separated from the pre-training and fine-tuning strategy, \cite{you2020does} has studied multi-task learning and analyzed why the pretext tasks are useful for GNNs.
However, one problem with both pre-training and multi-task learning strategies is that all the auxiliary tasks are not beneficial for the downstream applications.
So, we studied \textit{auxiliary learning} for GNNs that explicitly focuses on the primary task.
% To avoid the need for manual labeling, self-supervised learning on the target domain such as graph kernel~\cite{navarin2018pre}, graph reconstruction~\cite{zhang2020graph}, and attribute masking~\cite{hu2020strategies} has been proposed.
% Self-supervised learning on the target domain such as graph kernel~\cite{navarin2018pre}, graph reconstruction~\cite{zhang2020graph}, and attribute masking~\cite{hu2020strategies} has been proposed.
% The auxiliary tasks should be manually chosen with domain knowledge and they are not optimized for the primary task.

\textbf{Auxiliary Learning} is a learning strategy to employ auxiliary tasks to assist the primary task. 
It is similar to multi-task learning, but auxiliary learning cares only the performance of the primary task.
A number of auxiliary learning methods are proposed in a wide range of tasks~\cite{auxlearning1, Unaux1, Unaux2}. 
AC-GAN~\cite{acgan} proposed an auxiliary classifier for generative models.
Recently, Meta-Auxiliary Learning~\cite{maxl} proposes an elegant solution to generate new auxiliary tasks by collapsing existing classes. 
However, it cannot be applicable to some tasks such as link prediction which has only one positive class.
Our approach generates meta-paths on heterogeneous graphs to make new labels and trains models to predict meta-paths as auxiliary tasks.

\textbf{Meta-learning} aims at learning to learn models efficiently and effectively, and generalizes the learning strategy to new tasks. 
Meta-learning includes black-box methods to approximate gradients without any information about models~\cite{santoro2016meta, DBLP:conf/iclr/RaviL17},  
optimization-based methods to learn an optimal initialization for adapting new tasks~\cite{MAML, han2018coteaching, layerwise}, 
learning loss functions~\cite{han2018coteaching,learnwhere} and  
metric-learning or non-parametric methods for few-shot learning \cite{Koch2015SiameseNN, DBLP:conf/nips/SnellSZ17, DBLP:conf/cvpr/SungYZXTH18}.
In contrast to classical learning algorithms that generalize across samples, meta-learning generalizes across tasks.
% In contrast to classical learning algorithms that generalize the concept across samples, meta-learning generalizes across tasks.
In this paper, we use meta-learning to learn a concept across tasks and transfer the knowledge from auxiliary tasks to the primary task.

%% file: Sections/3_Method.tex
\section{Method} 
\label{headings}
The goal of our framework is to learn with multiple auxiliary tasks to improve the performance of the primary task. 
In this work, we demonstrate our framework with meta-path predictions as auxiliary tasks. But our framework could be extended to include other auxiliary tasks.
The meta-paths capture diverse and meaningful relations between nodes on heterogeneous graphs~\cite{HAN}.
However, learning with auxiliary tasks has multiple challenges: identifying useful auxiliary tasks, balancing the auxiliary tasks with the primary task, and converting challenging auxiliary tasks into solvable (and relevant) tasks. 
To address the challenges, we propose \textbf{SEL}f-supervised \textbf{A}uxiliary Lea\textbf{R}ning (\textbf{SELAR}). 
% First, useful auxiliary tasks to improve the primary task need to be identified.
% Even though the meta-path has rich semantic information, some meta-paths are not relevant to the primary task and it can damage the performance of primary task. 
% Further, learners should not be overwhelmed by the auxiliary tasks. Since the learners has the limited capacity, the primary and auxiliary tasks need to properly balanced. 
%% ***** Important message ******
%%  Comparison with Pre-training, We believe that Auxiliary Learning with meta-learning is a better approach than pre-training and 
%% fine-tuning !!!
% Why did we learn the sample wise weights?
% How about learning task-wise weights?
%
Our framework consists of two main components: 
% 1) learning weight functions to softly select auxiliary tasks (meta-paths) and balance them with the primary task (link) via meta-learning, and 
1) learning weight functions to softly select auxiliary tasks and balance them with the primary task via meta-learning, and
2) learning Hint Networks to convert challenging auxiliary tasks into more relevant and solvable tasks to the primary task learner.
\vspace{-0.15cm}
\subsection{Meta-path Prediction as a self-supervised task}
\input{Sections/3_1_Meta-path_Predictions.tex}
\subsection{Self-Supervised Auxiliary Learning}
\label{sec:selar}
\input{Sections/3_2_SELAR.tex}
\input{Sections/3_2_Learning_Algorithm.tex}

\vspace{-0.2cm}
\subsection{Hint Networks}
\label{sec:hintnet}
\input{Sections/3_4_Hint_Networks}

%% file: Sections/3_1_Meta-path_Predictions.tex
Most existing graph neural networks have been studied focusing on {\em homogeneous graphs} that have a single type of nodes and edges.
However, in real-world applications, {\em heterogeneous graphs}~\cite{heterogeneous}, which have multiple types of nodes and edges, commonly occur. 
Learning models on the heterogeneous graphs requires different considerations to effectively represent their node and edge heterogeneity. 

\textbf{Heterogeneous graph}~\cite{shi2016survey}. Let $G = (V,E)$ be a graph with a set of nodes $V$ and edges $E$. 
A heterogeneous graph  is a graph equipped with a node type mapping function $f_v : V \rightarrow \mathcal{T}^v$ and an edge type mapping function  $f_e : E \rightarrow \mathcal{T}^e$, where  $\mathcal{T}^v$ is a set of node types and $\mathcal{T}^e$ is a set of edge types. 
Each node $v_i \in V$ (and edge $e_{ij} \in E$ resp.) has one node type, i.e., $f_v(v_i) \in \mathcal{T}^v$, (and one edge type $f_e(e_{ij}) \in  \mathcal{T}^e$ resp.). In this paper, we consider the heterogeneous graphs with $|\mathcal{T}^e|> 1$ or $|\mathcal{T}^v|> 1$. When $|\mathcal{T}^e|=1$ and $|\mathcal{T}^v|=1$, it becomes a homogeneous graph.

\textbf{Meta-Path}~\cite{HAN,sun2011pathsim} is a path on a heterogeneous graph $G$ that a sequence of nodes connected with heterogeneous edges, i.e., ${v}_1 \xrightarrow{t_1} {v}_2 \xrightarrow{t_2} \ldots  \xrightarrow{t_l} {v}_{l+1}$, 
where $t_l \in \mathcal{T}^e$ denotes an $l$-th edge type of the meta-path. 
The meta-path can be viewed as a composite relation $R = t_1 \circ t_2 \ldots \circ t_l$ between node ${v}_1$ and ${v}_{l+1}$, where $R_1 \circ R_2$ denotes the composition of relation $R_1$ and $R_2$.
%The adjacency matrix based on meta-paths $\mathcal{P}$ with edge types $(t_1, t_2, \ldots, t_l)$ can be obtained by ${A}_\mathcal{P} = A_{t_l} \ldots  A_{t_2}A_{t_1}$, where $A_{t}$ is the adjacency matrix of edge type $t$.
The definition of meta-path generalizes multi-hop connections and is shown to be useful to analyze heterogeneous graphs.
For instance, in Book-Crossing dataset, `user-item-written.series-item-user' indicates that a meta-path that connects users who like the same book series.
%given two nodes $v_i$ and $v_j$, we can generate multiple different relations using meta-paths containing different intermediate nodes from each other, 
%\hjk{!!Let's find examples with looking more heterogeneous edges like the definition above.!!
%e.g., two music tracks can be connected via movies (two musics appearing in the same movie) and artists (two musics `sung by' the same artist).}

We introduce \textbf{meta-path prediction} as a self-supervised auxiliary task to improve the representational power of graph neural networks. 
To our knowledge, the meta-path prediction has not been studied in the context of self-supervised learning for graph neural networks in the literature.

% \textbf{Meta-path prediction} is similar to link prediction but meta-paths allow heterogeneous composite relations and multi-hop relations. 
\textbf{Meta-path prediction} is similar to link prediction but meta-paths allow heterogeneous composite relations. 
The meta-path prediction can be achieved in the same manner as link prediction. 
If two nodes $u$ and $v$ are connected by a meta-path $p$ with the heterogeneous edges $(t_1, t_2,\ldots t_\ell)$,  then $y_{u,v}^p=1$, otherwise $y_{u,v}^p=0$. The labels can be generated from a heterogeneous graph without any manual labeling. 
They can be obtained by $A_p = A_{t_l} \ldots  A_{t_2}A_{t_1}$, where $A_{t}$ is the adjacency matrix of edge type $t$. The binarized value at $(u, v)$ in $A_p$ indicates whether  $u$ and $v$ are connected with the meta-path $p$.
In this paper, we use meta-path prediction as a self-supervised auxiliary task. 

Let $\mathbf{X} \in \Rb^{|V| \times d }$ and $\mathbf{Z} \in \Rb^{|V| \times d'}$ be input features and their hidden representations learnt by GNN $f$, i.e., $\Zb = f(\Xb;\wb,\Ab)$, where $\wb$ is the parameter for $f$, and $\mathbf{A} \in \Rb^{|V|\times |V|}$ is the adjacency matrix. Then link prediction and meta-path prediction are obtained by a simple operation as
\begin{align}
\hat{y}_{u,v}^t = \sigma (\Phi_t(z_u)^\top \Phi_t(z_v) ),
\end{align}
where $\Phi_t$ is the task-specific network for task $t \in \Tc$ and $z_u$ and $z_v$ are the node embeddings of node $u$ and $v$. e.g., $\Phi_0$ (and $\Phi_1$ resp.) for link prediction (and the first type of meta-path prediction resp.). 

\begin{figure}
  \includegraphics[scale=0.43, bb=10 0 0 250]{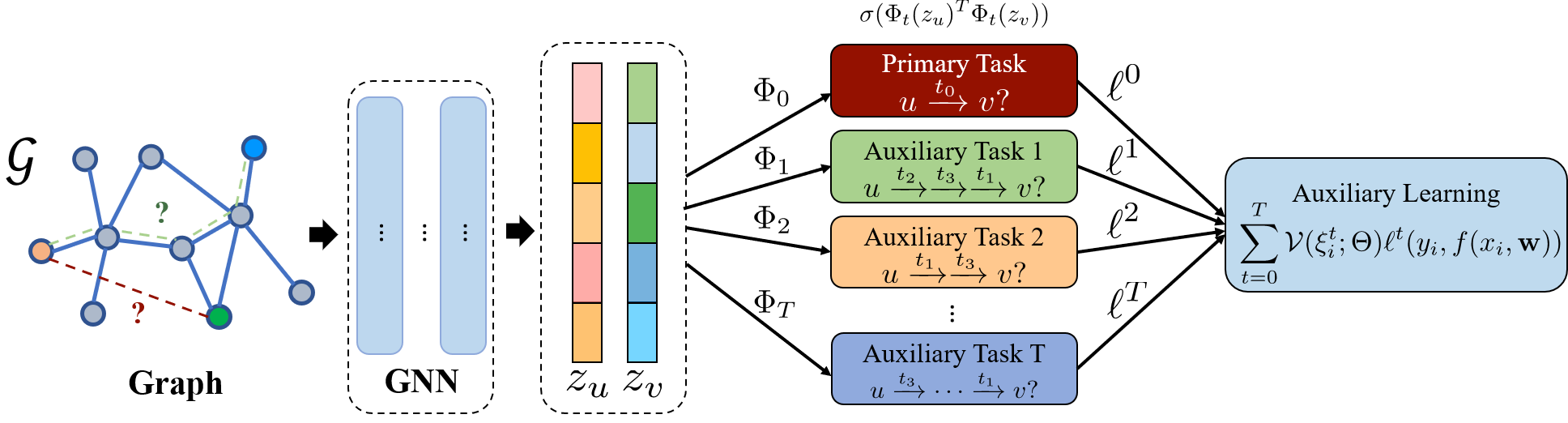}
  \caption{The SELAR framework for self-supervised auxiliary learning. Our framework learns how to balance (or softly select) auxiliary tasks to improve the primary task via meta-learning. In this paper, the primary task is link prediction (or node classification) and auxiliary tasks are meta-path predictions to capture rich information of a heterogeneous graph.}
  \label{fig:meta-path-prediction}
\end{figure}

\textbf{The architecture} is shown in Fig. ~\ref{fig:meta-path-prediction}.
To optimize the model, as the link prediction, cross entropy is used.
The graph neural network $f$ is shared by the link prediction and meta-path predictions. 
As any auxiliary learning methods, the meta-paths (auxiliary tasks) should be carefully chosen and properly weighted so that the meta-path prediction does not compete with link prediction especially when the capacity of GNNs is limited.
To address these issues, we propose our framework that automatically selects meta-paths and balances them with the link prediction via meta-learning.

%% file: Sections/3_2_SELAR.tex
Our framework SELAR is learning to learn a primary task with multiple auxiliary tasks to assist the primary task.
This can be formally written as 
\begin{align}
    &\min_{\wb,\Theta} \;\; \underset{(x,y) \sim D^{pr}\myrepeat{30}{\;}}{\text{\large $\EE$} \;\; \left [ \;\; \Lc^{pr}(\wb^{\ast}(\Theta))  \;\; \right ]} \;\; \text{ s.t. }\;\; \wb^{\ast}(\Theta) = \argmin_{\wb} \underset{(x,y) \sim D^{pr+au}\myrepeat{30}{\;}} {\;\; \EE \;\;\left [\;\;\Lc^{pr+au}(\wb;\Theta) \;\; \right ]},
    \label{eq:probsetup}
\end{align}
where $\Lc^{pr}(\cdot)$ is the primary task loss function to evaluate the trained model $f(x; \wb^{\ast}(\Theta))$ on meta-data (a validation for meta-learning~\cite{han2018coteaching}) $D^{pr}$ and $\Lc^{pr+au}$ is the loss function to train a model on training data $D^{pr+au}$ with the primary and auxiliary tasks. To avoid cluttered notation, $f$, $x$, and $y$ are omitted. Each task $\mathcal{T}_t$ has $N_t$ samples and $\Tc_0$ and $\{\Tc_t\}_{t=1}^T$ denote the primary and auxiliary tasks respectively. 
% Assume that the primary task (i.e., the $0_{th}$ task) has $N_0$ samples. That can be viewed as the meta-dataset for meta-learning. Each auxiliary task, $\mathcal{T}_t$ has $N_t$ training samples from training data. 
% In this paper, the training samples for the auxiliary tasks are generated by meta-paths without manual labeling. 
The proposed formulation in Eq.~\eqref{eq:probsetup} learns how to assist the primary task by optimizing $\Theta$ via meta-learning. The nested optimization problem given $\Theta$ is a regular training with properly adjusted loss functions to balance the primary and auxiliary tasks. The formulation can be more specifically written as 
\begin{align}
\min_{\wb, \Theta} & \sum_{i=1}^{M_0} \frac{1}{M_0}  \ell^0(y_i^{(0,meta)}, f(x_i^{(0,meta)};\wbst(\Theta)) \label{eq:Lmeta}\\ 
\text{s.t. } & \wbst(\Theta) = \argmin_{\wb} \sum_{t = 0}^T \sum_{i=1}^{N_t} \frac{1}{N_t} \Vc(\xi^{(t, train)}_i; \Theta) \ell^t(y_i^{(t,train)}, f^t(x_i^{(t,train)};\wb)) \label{eq:Ltrain},
\end{align}
where $\ell^t$ and $f^t$ denote the loss function and the model for task $t$. We overload $\ell^t$ with its function value, i.e., $\ell^t=\ell^t(y_i^{(t,train)}, f^t(x_i^{(t,train)};\wb))$. $\xi^{(t, train)}_i$ is the embedding vector of $i_{th}$ sample for task $t$. It is the concatenation of one-hot representation of task types, the label of the sample (positive/negative), and its loss value, i.e.,  $\xi^{(t, train)}_i = \left [\ell^t;e_t; y_i^{(t,train)} \right ] \in \Rb^{T+2}$.
%where $\ell^t$ is the loss function for task $t$ and $\wb$ is the model parameters, $f^t$ is a model for target $t$, the overloaded notation $\ell^t$ is used for the loss function and its function value, i.e., $\ell^t=\ell^t(y_i^{(t,train)}, f^t(x_i^{(t,train)};\wb)$, $\xi^{(t, train)}_i$ is the embedding vector of $i_{th}$ sample for task $t$. In our experiment, $\xi^{(t, train)}_i$ is the concatenation of one-hot representation of task types and the label of the sample, i.e.,  $\xi^{(t, train)}_i = \left [e_t; y_i^{(t,train)} \right ] \in \Rb^{T+2}$. \hjk{We will discuss the sample embedding in Sec. XXX.}

%% file: Sections/3_2_Learning_Algorithm.tex
To derive our learning algorithm, 
we first shorten the objective function in Eq.~\eqref{eq:Lmeta} and Eq.~\eqref{eq:Ltrain} as $\Lc^{pr}(\wbst(\Theta))$ and $\Lc^{pr+au}(\wb; \Theta)$.
This is equivalent to Eq.~\eqref{eq:probsetup} without expectation.
%we begin with the shortened form of our formulation in \eqref{eq:Lmeta} and \eqref{eq:Ltrain}.
%without the expectation for simplicity. 
Then, our formulation is given as 
\begin{equation}
\min_{\wb,\Theta}  \Lc^{pr}(\wbst(\Theta)) \;\; \text{ s.t. } \wbst(\Theta) = \argmin_{\wb} \Lc^{pr+au}(\wb; \Theta),
\label{eq:shortformulation}
\end{equation}
% To circumvent the difficulty of the bi-level optimization, we adopt the gradient update approximation as \cite{MAML, han2018coteaching}. 
%The formulation in \eqref{eq:Lmeta} and \eqref{eq:Ltrain} is a bi-level optimization problem. 
% To circumvent the difficulty of the bi-level optimization, we adopt the gradient update approximation as \cite{MAML, han2018coteaching}. 
% To introduce our learning algorithm, we first introduce the shorten form of our formulation.
% Let $\Lc^{pr}(\wbst(\Theta))$ and $\Lc^{pr+au}(\wb; \Theta)$ be the objective function in \eqref{eq:Lmeta} and \eqref{eq:Ltrain} respectively. 
% [After changing notations, it looks redundant. Needs polishing.] Then, our formulation can be re-written as
% \begin{equation}
% \min_{\wb,\Theta}  \Lc^{pr}(\wbst(\Theta)) \;\; \text{ s.t. } \wbst(\Theta) = \argmin_{\wb} \Lc^{pr+au}(\wb; \Theta).
% \label{eq:shortformulation}
% \end{equation}
% It is hard to obtain the optimal solution $\wbst(\Theta)$ to the nested optimization problem. 
% So 
To circumvent the difficulty of the bi-level optimization, as previous works~\cite{MAML, han2018coteaching} in meta-learning we approximate it with the updated parameters $\hat{\wb}$ using the gradient descent update as
% So we approximate the nested optimization by the updated parameters $\hat{\wb}$ using the gradient descent update as
\begin{align}
\wbst(\Theta) \approx \hat{\wb}^k(\Theta^k) = \wb^k - \alpha \nabla_{\wb} \Lc^{pr+au}(\wb^k; \Theta^k), \label{eq:w_hat}
\end{align}
where $\alpha$ is the learning rate for $\wb$. 
We do not numerically evaluate $\hat{\wb}^k(\Theta)$ instead we plug the computational graph of $\hat{\wb}^k$ in  $\Lc^{pr}(\wbst(\Theta))$ to optimize $\Theta$. 
Let $\nabla_{\Theta} \Lc^{pr}(\wbst(\Theta^k))$ be the gradient evaluated at $\Theta^k$.
Then updating parameters $\Theta$ is given as 
%done by the gradient descent with the intermediate variables $\hat{\wb}^k(\Theta)$ parameterized by $\Theta$, i.e., 
\begin{align}
\Theta^{k+1} = \Theta^{k} - \beta \nabla_{\Theta} \Lc^{pr}(\hat{\wb}^k(\Theta^k)), \label{eq:updateTheta}
\end{align}
where $\beta$ is the learning rate for $\Theta$. This update allows softly selecting useful auxiliary tasks (meta-paths) and balance them with the primary task to improve the performance of the primary task. Without balancing tasks with the weighting function $\Vc(\cdot ; \Theta)$, auxiliary tasks can dominate training and degrade the performance of the primary task.

The model parameters $\wb^k$ for tasks can be updated with optimized $\Theta^{k+1}$ in \eqref{eq:updateTheta} as
\begin{align}
\wb^{k+1} = \wb^{k} - \alpha \nabla_{\wb}\Lc^{pr+au} (\wb^k;\Theta^{k+1}). \label{eq:updateW}
\end{align}
\noindent\textbf{\it{Remarks.}} The proposed formulation can suffer from the meta-overfitting \cite{antoniou2018train,zintgraf2018fast} meaning that the parameters $\Theta$ to learn weights for softly selecting meta-paths and balancing the tasks with the primary task can overfit to the small meta-dataset. 
In our experiment, we found that the overfitting can be alleviated by meta-validation sets \cite{antoniou2018train}. 
To learn $\Theta$ that is generalizable across meta-training sets, we optimize $\Theta$ across $k$ different meta-datasets like $k$-fold  cross validation using the following equation:
\begin{align}
\Theta^{k+1} \myrepeat{1}{\;} = \myrepeat{1}{\;} \underset{D^{pr(meta)}\sim CV\myrepeat{
20}{\;}}{\Theta^{k} \myrepeat{1}{\;} - \myrepeat{2}{\;} \beta \; \;\EE \left [ \; \nabla_{\Theta} \Lc^{pr}(\hat{\wb}^k(\Theta^k))\; \right ],}\label{eq:updateThetaCV}
\end{align}
where $D^{pr(meta)}\sim CV$ is a meta-dataset from cross validation. We used 3-fold cross validation and the gradients of $\Theta$ w.r.t different meta-datasets are averaged to update $\Theta^k$, see Algorithm \ref{alg:selar}. The cross validation is crucial to alleviate meta-overfitting and more discussion is Section \ref{sec:exp_analysis}.

%\hjk{K-fold CV style meta-learning should be highlighted and please refer a paper for this.}
%\State $D^{train_s}, D^{meta_s}$ \leftarrow $CV(D^{pri}, S)$ \Comment{split data into $S$ folds} 
% \State $ \{x^{(t,train_s)},y^{(t,train_s)}\}, \{x^{(t,meta_s)},y^{(t,meta_s)}\} \leftarrow sampler(D^{train_s}, D^{meta_s})$
%			\State Sample a mini-batch of $D^{train_s}, D^{meta_s}$ and $D^{train}$ 
			%\State  $ \{x^{(0,meta_s)},y^{(0,meta_s)}\} \leftarrow Sampler(CV(D^{pri}, S))$
			%\State  ${\hat{\wb}_s}^k(\Theta) = \wb^k - \alpha \nabla_{\wb} \ell( y^{(t,train_s)},\Vc(\hat{\ell};\Theta) \cdot (\Phi_0(f) + \Phi_i(f;\Theta^\prime) )$
			%\State  ${\hat{\wb}_s}^k(\Theta) = \wb^k - \alpha \nabla_{\wb} \ell \left( y^{(t,train_s)},\Vc(\hat{\ell};\Theta) \cdot (\Phi_0(f) + \Phi_i(f;\Theta^\prime) \right)$
		%\State Sample a mini-batch of $D^{train_s}, D^{meta_s}$
		%\State Sample a mini-batch of $D^{train}$

	    % To do : S / T / CV / Theta / Thea_prime /

\input{Tables_tex/alg_selar}

%% file: Tables_tex/alg_selar.tex
\begin{algorithm}
\caption{\label{alg:selar}Self-supervised Auxiliary Learning} 
\textbf{Input:}  training data for primary/auxiliary tasks $D^{pr}, D^{au}$,  mini-batch size $N_{pr}, N_{au}$ \newline
\textbf{Input:}  max iterations $K$, \# folds for cross validation $C$, learning rate $\alpha, \beta$ \newline
\textbf{Output:}  network parameter $\wb^K$ for the primary task 
\setstretch{1.3}
%\setstretch{1.35}
\begin{algorithmic}[1]
\State Initialize $\wb^1, \Theta^1$ 
\For {$k=1$ to $K$}
\State $D^{pr}_{m} \leftarrow \text{MiniBatchSampler}(D^{pr}, N_{pr})$ 
\State $D^{au}_{m} \leftarrow \text{MiniBatchSampler}(D^{au}, N_{au})$ 
    \For {$c=1$ to $C$} \Comment{Meta Learning with Cross Validation}
    \State  $D^{pr(train)}_{m}, D^{pr(meta)}_{m} \leftarrow \text{CVSplit}(D^{pr}_m, c)$ 
    \Comment{Split Data for CV}
    \State  
    $\hat{\wb}^k(\Theta^k) \leftarrow \wb^k - \alpha \nabla_{\wb} \Lc^{pr+au}(\wb^k; \Theta^k)$ with $D^{pr(train)}_{m} \cup D^{au}_m$ \Comment{Eq. \eqref{eq:w_hat}}
    \State $g_c \leftarrow \nabla_{\Theta}\Lc^{pr}(\hat{\wb}^k(\Theta^k))$ with $D^{pr(meta)}_{m}$ \Comment{Eq. \eqref{eq:updateTheta}}
    \EndFor
\State Update $\Theta^{k+1} \leftarrow \Theta^{k} - \beta \sum_{c}^C g_c$ \Comment{Eq. \eqref{eq:updateThetaCV}} 
\State $\wb^{k+1} = \wb^{k} - \alpha \nabla_{\wb}\Lc^{pr+au} (\wb^k;\Theta^{k+1})$ with $D^{pr}_m \cup D^{au}_m$ \Comment{Eq. \eqref{eq:updateW}}
\EndFor
\end{algorithmic} 
\end{algorithm}

%% file: Sections/3_4_Hint_Networks.tex
\input{Figures_tex/fig_hint2}

Meta-path prediction is generally more challenging than link prediction and node classification since it requires the understanding of long-range relations across heterogeneous nodes.
The meta-path prediction gets more difficult when mini-batch training is inevitable due to the size of datasets or models.
Within a mini-batch, important nodes and edges for meta-paths are not available.
Also, a small learner network, e.g., two-layer GNNs, with a limited receptive field, inherently cannot capture long-range relations.
The challenges can hinder representation learning and damage the generalization of the primary task.
We proposed a Hint Network (HintNet) which makes the challenge tasks more solvable by correcting the answer with more information at the learner's need. 
Specifically, in our experiments, the HintNet corrects the answer of the learner with its own answer from the augmented graph with hub nodes, see Fig. ~\ref{fig:HintNet}. 

The amount of help (correction) by HintNet is optimized maximizing the learner's gain.
Let $\Vc_H(\cdot)$ and $\Theta_H$ be a weight function to determine the amount of hint and its parameters which are optimized by meta-learning. Then, our formulation with HintNet is given as 
\begin{align}
&\min_{\wb, \Theta}  \sum_{i=1}^{M_0} \frac{1}{M_0}  \ell^0(y_i^{(0,meta)}, f(x_i^{(0,meta)};\wbst(\Theta, \Theta_H))) \\ 
& \text{s.t. }  \wbst(\Theta) = \argmin_{\wb} \sum_{t = 0}^T \sum_{i=1}^{N_t} \frac{1}{N_t} \Vc(\xi^{(t, train)}_i, \ell^t; \Theta) \ell^t(y_i^{(t,train)}, \hat{y}_i^{(t,train)}(\Theta_{H})),
\end{align}
where $\hat{y}_i^{(t,train)}(\Theta_{H})$ denotes the convex combination of the learner's answer and HintNet's answer, i.e., $\Vc_H(\xi^{(t, train)}_i; \Theta_H) f^t(x_i^{(t,train)};\wb) + (1-\Vc_H(\xi^{(t, train)}_i; \Theta_H))f_H^t(x_i^{(t,train)};\wb)$. The sample embedding is
$\xi^{(t, train)}_i = \left [\ell^t; \ell^t_H; e_t; y_i^{(t,train)} \right ] \in \Rb^{T+3}$.

%% file: Figures_tex/fig_hint2.tex
\begin{wrapfigure}{tr}{0.5\textwidth}
\centering
\vspace{-0.9cm}
\includegraphics[width=0.5\textwidth]{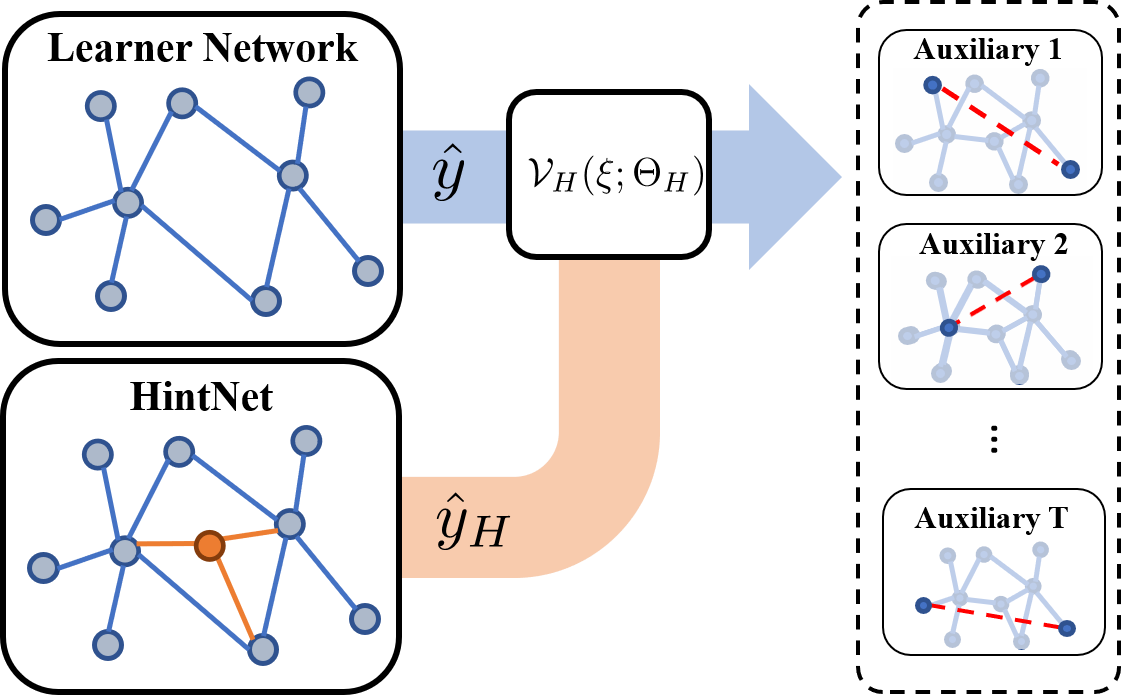}
\caption{HintNet helps the learner network to learn with challenging and remotely relevant auxiliary tasks. HintNet learns $\Vc_H$ to decide to use hint $\hat{y}_H$ in the \textcolor{orange}{orange} line or not. $\hat{y}$ in the \textcolor{blue}{blue} line denotes the prediction from the learner network.}
% \caption{HintNet learns $\Vc_H$ to decide to use hint $\hat{y}_H$ in the \textcolor{orange}{orange} line from HintNet or not via meta-learning. $\hat{y}$ in the \textcolor{blue}{blue} line denotes the prediction from the learner network.}
\label{fig:HintNet}
\vspace{-0.3cm}
\end{wrapfigure}

% \begin{wrapfigure}{tr}{0.5\textwidth}
% \centering
% \vspace{-1.198cm}
% \includegraphics[width=0.5\textwidth]{Figures/Figure_hint.png}
% \caption{HintNet helps the learner network to learn even with challenging and remotely relevant auxiliary tasks. As our framework selects effective auxiliary tasks, our framework with HintNet learns $\Vc_H$ to decide to use hint $\hat{y}_H$ in the \textcolor{orange}{orange} line from HintNet or not via meta-learning. $\hat{y}$ in the \textcolor{blue}{blue} line denotes the prediction from the learner network.}
% \label{fig:HintNet}
% \end{wrapfigure}

%% file: Sections/4_Experiments.tex
\section{Experiments}
\label{sec:experiments}
\input{Sections/4_1_Exp_Intro}

\subsection{Learning Link Prediction with meta-path prediction}
\input{Sections/4_2_Exp_Link_Prediction}
\subsection{Learning Node Classification with meta-path prediction}

\input{Sections/4_3_Exp_Node_Classification}
\subsection{Analysis of Weighting Function and Meta-overfitting}
\label{sec:exp_analysis}
\input{Sections/4_4_Weighting_Function_Analysis}

\input{Sections/4_5_Exp_Meta_Overfitting}

%% file: Sections/4_1_Exp_Intro.tex
We evaluate our proposed methods on four public benchmark datasets on heterogeneous graphs.
Our experiments answer the following research questions: \textbf{Q1.} Is meta-path prediction effective for representation learning on heterogeneous graphs?
\textbf{Q2.} Can the meta-path prediction be further improved by the proposed methods (e.g., SELAR, HintNet)?
\textbf{Q3.} Why are the proposed methods effective, any relation with hard negative mining?
%can we learn from the weight function $V(\xi_i;\Theta)$?
% the capability of the proposed method, and present its performance on heterogeneous graph. we implement experiments on four real-world datasets on two tasks such as node classification and link prediction.
%We aim to answer the following research questions:
%\begin{itemize}
%  \item Q1:
%  \item Q2:
%\end{itemize}
%\input{Tables_tex/tab_datasets}
%\subsection{Experimental Setup}

{\textbf{Datasets.} } 
We use two public benchmark datasets from different domains for link prediction: Music dataset Last-FM and Book dataset Book-Crossing, released by KGNN-LS~\cite{wang2019knowledge}, RippleNet~\cite{wang2018ripplenet}.
We use two datasets for node classification: citation network datasets ACM and Movie dataset IMDB, used by HAN~\cite{HAN} for node classification tasks.
% Last-FM contains listening history of two thousand users from Last.fm, an online music service.
% Book-Crossing contains 278,858 users providing 1,149,780 explicit ratings of books in the Book-crossing community. 
ACM has three types nodes (Paper(P), Author(A), Subject(S)), four types of edges (PA, AP, PS, SP) and labels (categories of papers). 
% In our experiments, we use meta-path sets PAP, PSP for performing experiments.
IMDB contains three types of nodes (Movie (M), Actor (A), Director (D)),  four types (MA, AM, MD, DM) of edges  and labels (genres of movies). 
%We use meta-path sets MAM, MDM for performing experiments.
ACM and IMDB have node features, which are bag-of-words of keywords and plots.
Dataset details are in the supplement. 

{\textbf{Baselines.} }
We evaluate our methods with five graph neural networks : GCN~\cite{GCN}, GAT~\cite{GAT}, GIN~\cite{xu2018powerful}, SGConv~\cite{wu2019simplifying} and GTN~\cite{yun2019graph}. Our methods can be applied to both homogeneous graphs and heterogeneous graphs. We compare four learning strategies: \textbf{Vanilla}, standard training of base models only with the primary task samples; \textbf{w/o meta-path}, learning a primary task with sample weighting function $\Vc(\xi;\Theta)$; \textbf{w/ meta-path}, training with the primary task and auxiliary tasks (meta-path prediction) with a standard loss function; \textbf{SELAR} proposed in Section \ref{sec:selar}, learning the primary task with optimized auxiliary tasks by meta-learning; \textbf{SELAR+Hint} introduced in Section \ref{sec:hintnet}.
In all the experiments, we report the mean performance of three independent runs.
% In all the experiments, the evaluations are conducted with three independent runs.
% In all the experiments, we evaluate models three independent runs.
Implementation details are in the supplement. Our experiments were mainly performed based on NAVER Smart Machine Learning platform (NSML)~\cite{sung2017nsml,kim2018nsml}.

%% file: Sections/4_2_Exp_Link_Prediction.tex
We used five types of meta-paths of length 2 to 4 for auxiliary tasks.
Table~\ref{tab:link prediction} shows that our methods consistently improve link prediction performance for all the GNNs, compared to the Vanilla and the method using Meta-Weight-Net~\cite{han2018coteaching} only without meta-paths (denoted as w/o meta-path). 
Overall, a standard training with meta-paths shows 1.1\% improvement on average on both Last-FM and Book-Crossing whereas meta-learning that learns sample weights degrades on average on Last-FM and improves only 0.6\% on average on Book-Crossing, e.g., GCN, SGC and GTN on Last-FM and GCN and SGC on Book-Crossing, show degradation 0.2\% compared to the standard training (Vanilla). 
As we expected, SELAR and SELAR with HintNet provide more optimized auxiliary learning resulting in  1.9\% and 2.0\% absolute improvement on Last-FM and 2.6\% and 2.7\% on the Book-Crossing dataset.
Further, in particular, GIN on Book-crossing, SELAR and SELAR+Hint provide $\sim$5.5\% and $\sim$5.3\% absolute improvement compared to the vanilla algorithm.

\input{Tables_tex/tab_link_prediction}

%% file: Tables_tex/tab_link_prediction.tex
\setlength{\tabcolsep}{8pt}
\begin{table}[ht]
  \centering
  \caption{\textbf{Link prediction} performance ($AUC$) of GNNs trained by various learning strategies.}
  \footnotesize 
  \label{tab:link prediction}
  \begin{tabular}{c@{ }c|rr|rrr}
    \toprule
    \multirow{2}*{Dataset}&\multirow{2}*{Base GNNs}&\multirow{2}*{Vanilla}&\multirow{2}{*}{\shortstack{w/o \\  meta-path}} & \multicolumn{3}{c}{Ours} \\
    %\hline
    & & & & w/ meta-path & SELAR & SELAR+Hint \\
    %\toprule
    \midrule
    \multirow{6}{*}{Last-FM} 
    & GCN & 0.7963  & 0.7889 & 0.8235 & \textbf{0.8296} & 0.8121  \\
    & GAT & 0.8115  & 0.8115 & 0.8263 & 0.8294 & \textbf{0.8302}\\
    & GIN & 0.8199  & 0.8217 & 0.8242 & \textbf{0.8361} & 0.8350\\
    & SGC & 0.7703  & 0.7766 & 0.7718 & 0.7827 & \textbf{0.7975} \\
    & GTN & 0.7836  & 0.7744 & 0.7865 & 0.7988 & \textbf{0.8067} \\
    \cmidrule{2-7}
    & Avg. Gain & - & -0.0017 & +0.0106 & +0.0190 & +0.0200 \\
    \midrule
    \multirow{6}{*}{Book-Crossing} 
    & GCN & 0.7039 & 0.7031 & 0.7110 & 0.7182 & \textbf{0.7208}\\
    & GAT & 0.6891 & 0.6968  & 0.7075 & 0.7345 & \textbf{0.7360} \\
    & GIN & 0.6979 & 0.7210  & 0.7338 & \textbf{0.7526} & 0.7513\\
    & SGC & 0.6860 & 0.6808 & 0.6792 & 0.6902 & \textbf{0.6926} \\
    & GTN & 0.6732 & 0.6758 & 0.6724 & \textbf{0.6858} & 0.6850 \\
    \cmidrule{2-7}
    & Avg. Gain & - & +0.0055 & +0.0108 & +0.0263 & +0.0267\\
    \bottomrule
  \end{tabular}
\end{table}

%% file: Sections/4_3_Exp_Node_Classification.tex
\input{Tables_tex/tab_node_classification}

%\emph{\textbf{Node classification.}}
Similar to link prediction above, our SELAR consistently enhances node classification performance of all the  GNN models and the improvements are more significant on IMDB which is larger than the ACM dataset. 
We believe that ACM dataset is already saturated and the room for improvement is limited. However, our methods still show small yet consistent improvement over all the architecture on ACM. 
We conjecture that the efficacy of our proposed methods differs depending on graph structures. 
However, it is worth noting that introducing meta-path prediction as auxiliary tasks remarkably improves the performance of primary tasks such as link and node prediction with consistency compared to the existing methods. ``w/o meta-path'', the meta-learning to learn sample weight function on a primary task shows marginal degradation in five out of eight settings.
Remarkably, SELAR improved the F1-score of GAT on the IMDB by (4.46\%) compared to the vanilla learning scheme.

% Backup
% The results of node classification are almost consistent to ones of link prediction. Table~\ref{tab:node classification} depicts node classification results on the ACM and the IMDB datasets with three GNNs including GCN, GAT, and SGC. Same as Table~\ref{tab:link prediction}, our Selar consistently enhances node classification peformances of all the base GNNs and the improvements are more remarkable on IMDB which is larger than the ACM dataset. Similar to link prediction, the effectiveness of each component of our method depends on the used GNNs and datasets. We conjecture that these efficacy differences of our Selar components might be determined by the given graph properties. However, it is worth noting that introducing meta-path prediction as auxiliary tasks remarkably improve the performance of primary tasks such as link and node prediction with consistency compared to the existing methods.

%% file: Tables_tex/tab_node_classification.tex
\begin{table}[ht]
  \centering
  \caption{\textbf{Node classification} performance ($F1$-score) of GNNs trained by various learning schemes.}
  \footnotesize
  \label{tab:node classification}
  \begin{tabular}{cc|rr|rrr}
    \toprule
    \multirow{2}*{Dataset}&\multirow{2}*{Base GNNs}&\multirow{2}*{Vanilla}&\multirow{2}{*}{\shortstack{w/o \\  meta-path}} & \multicolumn{3}{c}{Ours} \\
    & & & & w/ meta-path & SELAR & SELAR+Hint \\
    \midrule
    \multirow{6}{*}{ACM} & GCN & 0.9091 & 0.9191 & 0.9104 & 0.9229 & \textbf{0.9246} \\
    & GAT & 0.9161 & 0.9119 & 0.9262 & 0.9273 & \textbf{0.9278} \\
    & GIN & 0.9085 & 0.9118 & 0.9058 & 0.9092 & \textbf{0.9135}\\
    & SGC & 0.9163 & 0.9194 & 0.9223 & 0.9224 & \textbf{0.9235} \\
    & GTN & 0.9181 & 0.9191 & 0.9246 & \textbf{0.9258}  &  0.9236\\
    \cmidrule{2-7}
    & Avg. Gain & - & +0.0027 & +0.0043 & +0.0079 & \textbf{+0.0090} \\
    \midrule
    \multirow{6}{*}{IMDB} & GCN & 0.5767 & 0.5855 & 0.5994 & 0.6083 & \textbf{0.6154}\\
    & GAT & 0.5653 & 0.5488 & 0.5910 & \textbf{0.6099} & 0.6044 \\
    & GIN & 0.5888 & 0.5698 & 0.5891 & \textbf{0.5931} & 0.5897 \\
    & SGC & 0.5779 & 0.5924 & 0.5940 & 0.6151 & \textbf{0.6192} \\
    & GTN & 0.5804 & 0.5792 & 0.5818 & 0.5994 & \textbf{0.6063} \\
    \cmidrule{2-7}
    & Avg. Gain & - & -0.0027 & +0.0132 & +0.0274 & \textbf{+0.0292} \\
    \bottomrule
  \end{tabular}
\end{table}

% \begin{table}[h]
%   \centering
%   \caption{Node classification performance ($F1$-score) of GNNs trained by various learning schemes.}
%   \label{tab:node classification}
%   \begin{tabular}{c|c|rr|rrr}
% %    \toprule
% %    Dataset & Basemodel & Vanilla & w/o meta-path & w/ meta-path & Selar & Selar+Hint \\
% %    \toprule
%     %\toprule
%     \hline\hline
%     \multirow{2}*{Dataset}&\multirow{2}*{Base GNNs}&\multirow{2}*{Vanilla}&\multirow{2}*{w/o meta-path} & \multicolumn{3}{c}{Ours} \\
%     %\hline
%     & & & & w/ meta-path & SELAR & SELAR+Hint \\
%     %\toprule
%     \hline
%     \multirow{5}{*}{ACM} & GCN & 0.9034 & $^*$0.9025 & 0.9147 & 0.9031 & \textbf{0.9160} \\
%     & GAT & 0.9179 & 0.9092 & 0.9188 & \textbf{0.9198} & 0.9188 \\
%     & GIN & 0.9060 & 0.9130 & 0.9101 & 0.9076 & \textbf{0.9135}\\
%     & SGC & 0.9138 & $^*$0.9115 & \textbf{0.9202} & 0.9120 & 0.9171 \\
%     \cline{2-7}
%     & Avg. Gain & - & -0.0013 & +0.0057 & +0.0003 & \textbf{+0.0061} \\
%     %\midrule
%     \hline
%     \multirow{5}{*}{IMDB} & GCN & 0.5826 & 0.5952 & \textbf{0.6189} & 0.6072 & 0.5970\\
%     & GAT & 0.5587 & $^*$0.5543 & 0.6013 & \textbf{0.6197} & 0.6017 \\
%     & GIN & 0.5965 & $^*$0.5856 & 0.5974 & \textbf{0.5994} & $^*$0.5726 \\
%     & SGC & 0.6163 & $^*$0.6156 & $^*$0.5966 & \textbf{0.6224} & $^*$0.5906 \\
%     \cline{2-7}
%     & Avg. Gain & - & -0.0008 & +0.0156 & \textbf{+0.0237} & +0.0020 \\
%     %\bottomrule
%     \hline\hline
%   \end{tabular}
% \end{table} 

%% file: Sections/4_4_Weighting_Function_Analysis.tex
The effectiveness of meta-path prediction and the proposed learning strategies are answered above. 
To address the last research question \textbf{Q3.} why the proposed method is effective, we provide analysis on the weighting function $\Vc(\xi;\Theta)$ learned by our framework. Also, we show the evidence that meta-overfitting occurs and can be addressed by cross-validation as in Algorithm \ref{alg:selar}.

\input{Figures_tex/fig_weight}

\textbf{Weighting function.} Our proposed methods can automatically balance multiple auxiliary tasks to improve the primary task. 
To understand the ability of our method, we analyze the weighting function and the adjusted loss function by the weighting function, i.e.,$\Vc(\xi;\Theta)$, $\Vc(\xi;\Theta)\ell^t(y,\hat{y})$. 
The positive and negative samples are solid and dash lines respectively.
We present the weighting function learnt by SELAR+HintNet for GAT which is the best-performing construction on Last-FM. The weighting function is from the epoch with the best validation performance.
Fig.~\ref{fig:weighfunc} shows that the learnt weighting function attends to hard examples more than easy ones with a small loss range from 0 to 1. 

Also, the primary task-positive samples are relatively less down weighted than auxiliary tasks even when the samples are easy (i.e., the loss is ranged from 0 to 1). Our adjusted loss $\Vc(\xi;\Theta)\ell^t(y,\hat{y})$ is closely related to the focal loss, $-(1-p_t)^{\gamma}\log(p_t)$. When $\ell^t$ is the cross-entropy, it becomes $\Vc(\xi;\Theta)\log(p_t)$, where $p$ is the model's prediction for the correct class and $p_t$ is defined as $p$ if $y=1$, otherwise $1-p$ as \cite{lin2017focal}.
The weighting function differentially evolves over iterations. At the early stage of training, it often focuses on easy examples first and then changes its focus over time. Also, the adjusted loss values by the weighting function learnt by our method differ across tasks. To analyze the contribution of each task, we calculate the average of the task-specific weighted loss on the Last-FM and Book-Crossing datasets. Especially, on the Book-Crossing, our method has more attention to 'user-item' (primary task) and `user-item-literary.series.item-user' (auxiliary task) which is a meta-path that connects users who like a book series. This implies that two users who like a book series likely have a similar preference.
% who like a book series have similar preference. 
More results and discussion are available in the supplement.

%% file: Figures_tex/fig_weight.tex
\begin{figure}[ht]
\centering
\begin{subfigure}{0.4\textwidth}
    \centering
    \hspace{-1.2cm}
	\includegraphics[height=4cm]{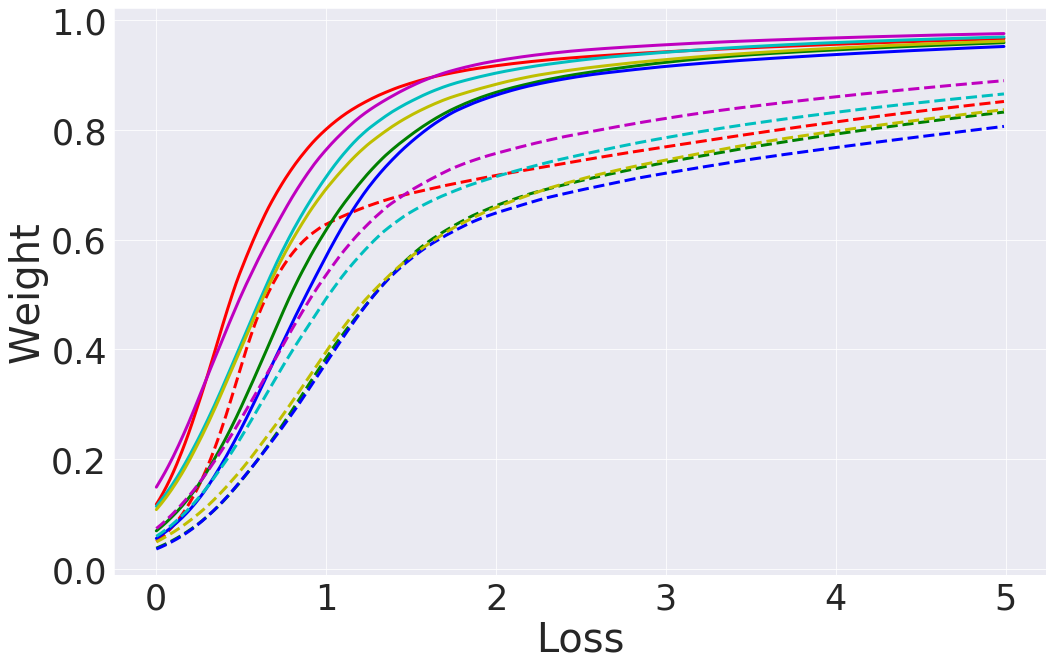}
	\caption{Weighting function $\Vc(\xi;\Theta)$.            }
	\label{fig:weighfunc_a}
\end{subfigure}
\begin{subfigure}{0.4\textwidth}
    \hspace{-0.5cm}
	\includegraphics[height=4cm]{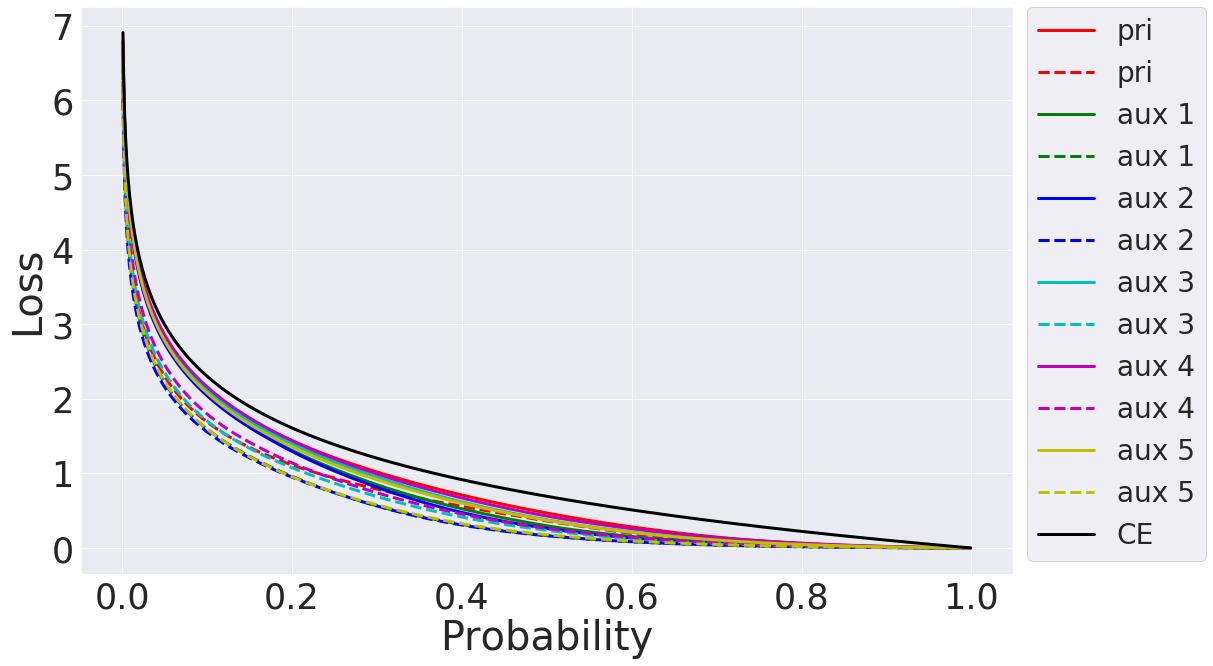}
	\caption{Adjusted Cross Entropy $\Vc(\xi;\Theta)\ell^t(y,\hat{y})$.}
	\label{fig:weighfunc_b}
\end{subfigure}
\caption{Weighting function $\Vc(\cdot)$ learnt by SELAR+HintNet. $\Vc(\cdot)$ gives overall high weights to the primary task positive samples (\textcolor{red}{red}) in (a). $\Vc(\cdot)$ decreases the weights of easy samples with a loss ranged from 0 to 1. In (b), the adjusted cross entropy, i.e., $-\Vc(\xi;\Theta)\log(\hat{y})$, by $\Vc(\cdot)$ acts like the focal loss, which focuses on hard examples by $-(1-p_t)^{\gamma}\log(\hat{y})$.
% $\Vc(\cdot)$ adjusts the cross entropy (CE), \textcolor{black}{black} line in (b), as $-\Vc(\xi;\Theta)\log(\hat{y})$, which is similar to the focal loss. The adjusted loss focuses on hard samples.
}
\label{fig:weighfunc}
\end{figure}

	%\includegraphics[trim=0 0 210 0, clip,height=4cm]{Figures/loss-best.png}%
	%\includegraphics[trim=1010 0 0 0, clip,height=4cm]{Figures/loss-best.png}

% Weighting function learned in Last-FM.

%\begin{figure}
	%\centering
	%\begin{subfigure}{0.99\textwidth} 
		%\includegraphics[trim=0 0 0 0, clip, width=0.45\textwidth]{Figures/1-weight-epoch1.png}%
		%\includegraphics[trim=0 0 260 0, clip, width=0.45\textwidth]{Figures/1-weight-epoch62.png}%
		%\includegraphics[trim=0 0 250 0, clip, width=0.28\textwidth]{Figures/1-weight-epoch100.png}%
%		\includegraphics[trim=1030 0 0 0, clip, height=4cm]{Figures/1-weight-epoch100.png}
%	\end{subfigure}
%\end{figure}

%\begin{itemize}
%  \item 모델 4개 모두 처음에는 샘플마다 비슷한 weight를 갖음 (y-scale 확인)
%  \item GAT의 경우 두번째 열(Best performance epoch)에서 잘 맞추는 샘플들에 대한 가중치는 낮추고, 못맞추는 샘플들에 대한 가중치를 높임.
%\end{itemize}

%\begin{figure}[h]
%    \centering
%    \includegraphics[width=1\textwidth]{Figures/weight-gcn-gat.png}
%    \caption{Draft: weight function in Last-FM \\
%    x-axis: loss, y-axis: weight }
%\label{fig:Weight}
%\end{figure}

%\begin{figure}[h]
%    \centering
%    \includegraphics[width=1\textwidth]{Figures/best-im-all.png}
%    \caption{Draft: weight function in Last-FM \\
%    x-axis: loss, y-axis: weight }
%\label{fig:Weight}
%\end{figure}

%% file: Sections/4_5_Exp_Meta_Overfitting.tex
\textbf{Meta cross-validation}, i.e., cross-validation for meta-learning, helps to keep weighting function from  over-fitting on meta data. 
Table~\ref{tab:metacv_edge} evidence that our algorithms as other meta-learning methods can overfit to meta-data. 
As in Algorithm \ref{alg:selar}, our proposed methods, both SELAR and SELAR with HintNet,  with cross-validation denoted as `3-fold' alleviates the meta-overfitting problem and provides a significant performance gain, whereas without meta cross-validation denoted as `1-fold' the proposed method can underperform the vanilla training strategy.

\input{Tables_tex/tab_CV_LastFM.tex}

%% file: Tables_tex/tab_CV_LastFM.tex
\begin{table}[h]
  \caption{Comparison between 1-fold and 3-fold as meta-data on \textbf{Last-FM} datasets.}
  \label{tab:metacv_edge}
  \centering
  \begin{tabular}{c|c|cc|cc}
    \toprule
     & & \multicolumn{2}{c}{SELAR} & \multicolumn{2}{c}{SELAR+Hint}\\
     
     Model &Vanilla &1-fold & 3-fold & 1-fold & 3-fold  \\
    \midrule
     GCN & 0.7963 & 0.7885 & \textbf{0.8296} & 0.7834 & \textbf{0.8121}  \\
     GAT & 0.8115 & 0.8287 & \textbf{0.8294} & 0.8290 & \textbf{0.8302}  \\
     GIN & 0.8199 & 0.8234 & \textbf{0.8361} & 0.8244 & \textbf{0.8350}  \\
     SGC & 0.7703 & 0.7691 & \textbf{0.7827} & 0.7702 & \textbf{0.7975}  \\
     GTN & 0.7836 & 0.7897 & \textbf{0.7988} & 0.7915 & \textbf{0.8067}  \\
    \bottomrule
  \end{tabular}
\end{table}
% \vspace{-1cm}

%% file: Sections/5_Conclusion.tex
\section{Conclusion}
\label{sec:conclusion}
We proposed meta-path prediction as self-supervised auxiliary tasks on heterogeneous graphs. 
Our experiments show that the representation learning on heterogeneous graphs
can benefit from meta-path prediction which encourages to capture rich semantic information.
%of the complex graph structure.
The auxiliary tasks can be further improved by our proposed method SELAR, which automatically balances auxiliary tasks to assist the primary task via a form of meta-learning. 
The learnt weighting function identifies more beneficial meta-paths for the primary tasks. 
Within a task, the weighting function can adjust the cross entropy like %better wording
the focal loss, which focuses on hard examples by decreasing weights for easy samples. 
Moreover, when it comes to challenging and remotely relevant auxiliary tasks, 
our HintNet helps the learner by correcting the learner's answer dynamically and further improves the gain from auxiliary tasks.
Our framework based on meta-learning provides learning strategies to balance primary task and auxiliary tasks, and easy/hard (and positive/negative) samples. 
Interesting future directions include applying our framework to other domains and various auxiliary tasks.
Our code is publicly available at \url{https://github.com/mlvlab/SELAR}.

%% file: Sections/6_Acknowledgements.tex
\noindent\textbf{Acknowledgements.}\quad
%%%%% MLV %%%%%
This work was partly supported by NAVER Corp. and Institute for Information \& communications Technology Planning \& Evaluation (IITP) grants funded by the Korea government (MSIT): the Regional Strategic Industry Convergence Security Core Talent Training Business (No.2019-0-01343) and the ICT Creative Consilience Program (IITP-2020-0-01819).

% This work was partially supported by NAVER Corp., Regional strategic Industry convergence security core talent training business (No.2019-0-01343, Regional strategic Industry convergence security core talent training business) and the ICT Creative Consilience program (IITP-2020-0-01819) supervised by the IITP (Institute for Information \& communications Technology Planning \& Evaluation).

% This work was partially supported by NAVER Corp., Institute for Information \& communications Technology Planning \& Evaluation (IITP) grant funded by the Korea government (MSIT) (No.2019-0-01343, Regional strategic Industry convergence security core talent training business) and MSIT (Ministry of Science and ICT), Korea, under the ICT Creative Consilience program (IITP-2020-0-01819) supervised by the IITP(Institute for Information \& communications Technology Planning \& Evaluation).

% This research was supported by the MSIT (Ministry of Science and ICT), Korea, under the ICT Creative Consilience program (IITP-2020-0-01819) supervised by the IITP(Institute for Information & communications Technology Planning & Evaluation).

%% file: Sections/BroaderImpact.tex
\section*{Broader Impact}
We thank NeurIPS2020 for this opportunity to revisit the broader impact of our work and the potential societal consequence of machine learning researches. Our work is a general learning method to benefit from auxiliary tasks. One interesting finding is that meta-path prediction can be an effective self-supervised task to learn more power representation of heterogeneous graphs.  Nowadays, people use social media (e.g., Facebook, Twitter, etc.) on a daily basis.  Also, people watch movies and TV-shows online and purchase products on Amazon. All this information can be represented as heterogeneous graphs. We believe that our meta-path auxiliary tasks will benefit the customers with improved services. 
For instance, more accurate recommender systems will save customers' time and provide more relevant contents and products.  We believe that there is no direct negative consequence of this research. We proposed how to train models with auxiliary tasks. We did not make any algorithms for specific applications. So, no one will be put at a disadvantage from our work. No direct negative consequence of a failure of the system is expected. We used four datasets Last-FM, Book-Crossing, ACM, and IMDB. They may not represent all the population on the earth but our experiments did not leverage any biases in the datasets. We believe that our method will be as effective as we reported in the paper on different datasets from different populations.

% The graph neural networks learnt by our proposed methods are not directly  since
% In terms of expected disadvantage from this research
% Authors are required to include a statement of the broader impact of their work, including its ethical
% aspects and future societal consequences. Authors should discuss both positive and negative outcomes,
% if any. For instance, authors should discuss a) who may benefit from this research, b) who may be
% put at disadvantage from this research, c) what are the consequences of failure of the system, and d)
% whether the task/method leverages biases in the data. If authors believe this is not applicable to them,
% authors can simply state this.
% Use unnumbered first level headi